# LAVOLUTION: Measurement of Non-target Structural Displacement Calibrated by Structured Light

Jongbin Won[a], Minhyuk Song[a], Gunhee Kim[a], Jong-Woong Park[a*] and Haemin Jeon[b]

[a]Department of Civil and Environmental Engineering, Chung-Ang University, Dongjak, Seoul 06974, Korea
[b]Department of Civil and Environmental Engineering, Hanbat National University, Yusung, Daejeon 34158, Korea

* Corresponding author: Tel: +82-3167-3275, E-mail address: jongwoong@cau.ac.kr

*Abstract—*

Displacement is an important measurement for the assessment of structural conditions, but its field measurement is often hindered by difficulties associated with sensor installation and measurement accuracy. To overcome the disadvantages of conventional displacement measurement, computer vision (CV)-based methods have been implemented due to their remote sensing capabilities and accuracy. This paper presents a strategy for non-target structural displacement measurement that makes use of CV to avoid the need to install a target on the structure while calibrating the displacement using structured light. The proposed system called as LAVOLUTION calculates the relative position of the camera with regard to the structure using four equally spaced beams of structured light and obtains a scale factor to convert pixel movement into structural displacement. A jig for the four beams of structured light is designed and a corresponding alignment process is proposed. A method for calculating the scale factor using the designed jig for tunable structured-light is proposed and validated via numerical simulations and lab-scale experiments. To confirm the feasibility of the proposed displacement measurement process, experiments on a shaking table and a full-scale bridge are conducted and the accuracy of the proposed method is compared with that of a reference laser doppler vibrometer.

# 1. Introduction

Structural health monitoring (SHM) has emerged as an essential process for ensuring the safety and serviceability of a structure and is particularly important for damage detection and failure prevention during operation. In the monitoring of a structure, structural responses to operational and environmental loads are measured, with displacement one of the most important of these responses for indicating the condition of a structure. Displacement can be measured using contact or non-contact methods. Traditional contact approaches include high-precision displacement transducers, such as linear variable differential transformers (LVDTs). Although LVDTs have the advantage of enabling high-precision displacement measurements, they depend on a fixed reference point, limiting their implementation in many cases (e.g., structures located over waterways and roads) (Gomez et al., 2018; Sarwar & Park, 2020; Won et al., 2021). Non-contact measurement methods based on laser, radar, GPS, and computer vision have been studied because they do not require a fixed reference point. Laser doppler vibrometers (LDVs) can measure high-resolution displacement, but they are expensive and can only measure the displacement in the direction of the emitted laser. An indirect displacement measurement method using the global positioning system (GPS) has been proposed; however, though it has an accuracy that makes it suitable for cases of significant displacement, it cannot handle precise measurements on a millimeter scale.

Recently, computer vision (CV)-based methods have become a useful approach to the indirect measurement of structural displacement because of their cost-effectiveness, high resolution, and relatively simple implementation (Bhowmick et al., 2020; Dong & Catbas, 2021; D. Feng & Feng, 2017; M. Q. Feng et al., 2015; Hoskere et al., 2019; J. Lee et al., 2020; Lydon et al., 2019; Song et al., 2022; Tian et al., 2019; Weng et al., 2021; Xu & Brownjohn, 2018; Yoon et al., 2016; Zhang et al., 2021). CV-based displacement measurement can be divided into two systems: target- and non-target-based. Target-based methods use artificial targets and tracking reference points, such as transition points between different colors (G. Lee et al., 2022; Luo et al., 2018; Ribeiro et al., 2021). For example, targets consisting of high-contrast black and white squares (e.g., a checkerboard pattern and ArUco marker; Garrido-Jurado et al., 2016; Romero-Ramirez et al., 2018) allow for accurate displacement measurement using various tracking algorithms. However, physical access to the surface of the structure is required to attach the target, which can be challenging or impossible in many cases. To address this limitation of target-based methods, non-target-based methods have been developed to employ natural features on a structure as targets (Dong & Catbas, 2019; Jeong & Jo, 2022; Khuc & Catbas, 2017; S. Lee et al., 2020; Narazaki et al., 2021; Won et al., 2019; Zhu et al., 2021). Because non-target-based methods do not have artificial references that can be used to convert camera coordinates into real-world coordinates, they generally track regions that contain distinct features such as stud bolts rather than the flat surface of a structure. However, in most cases, image acquisition for displacement measurement is taken at an inclined or tilted angle due to the camera installation environment. If feature points protruding from the structure, such as stud bolts, are tracked at a tilted angle, it is difficult to accurately scale them pixel-wise to real-world displacement. Hence, in order to obtain accurate displacement measurements, a method that is capable of accurate scaling regardless of the angle at which an image or a video of the structure is taken is required.

Structured-light (SL) systems have been explored for vision-based inspection because of their lower time and economic costs (Andrews, 2011; Chen et al., 2008; Fofi et al., 2004; Kawasaki et al., 2008; McPherron et al., 2009; Salvi et al., 2010; Scharstein & Szeliski, 2003; Zexiao et al., 2005). An SL system usually consists of a projector and a camera (also referred to as the transmitter and the receiver, respectively). A pattern composed of points and lines is emitted by the projector onto the object, and the camera receives the deformed pattern. Myung et al. (2011) developed a multiple-pair SL system to measure the displacement of a long-span structure. One unit was composed of two modules facing each other, each with two laser beams, a screen, and a camera. The system could measure displacement with six degrees of freedom based on the projection of the laser beams from one side onto the screen on the other side. Jeon et al. (2017) also developed a visually servoed paired SL system (ViSP) consisting of a transmitter and a receiver. The transmitter consisted of a laser range finder, two lasers, and a manipulator that emits the laser beams, with a camera fixed on the receiver detecting the projected beams on a screen. Six-degrees-of-freedom displacement can be estimated based on the coordinates of the beams, the distance obtained from a laser range finder, and the rotation angle of the manipulator. However, the installation of the receiver on a structure can be limited due to the same problem as the contact method or the target-based method.

To overcome the drawbacks of paired SL systems, this study proposes a single integrated SL system composed of an SL scaler and a camera. The SL scaler consists of four laser beams and a laser distance meter, which is used to obtain the calibration parameters. Because the four laser points projected onto a structure are captured by a camera within a single module, there is no need to calibrate the geometry of the lasers and the camera. The distance information obtained from the laser distance meter is used to calculate the scaling factor to calibrate the structural displacement.

The remainder of this paper is organized as follows. Section 2 summarizes the development and alignment of the jig, which is composed of a camera and tunable SL. It also describes the calibration of the camera rotation in order to calculate and monitor the scale factor for the measurement of the structural displacement. Section 3 describes the lab-scale experiment used to validate the proposed method, while Section 4 summarizes the experimental results using a real bridge to confirm the feasibility of the proposed method. Finally, an overall summary and conclusion are presented in Section 5.

## 2. Proposed method

*2.1. Overview*

The proposed method consists of three steps: (1) alignment of jig of LAVOLUTION to ensure that the optical axis and lasers are parallel; (2) the calculation of the scale factor to convert pixel coordinates to real-world coordinates; and (3) structural displacement measurement using the optical flow and calculated scale factor. The jig of LAVOLUTION contains a camera, four laser pointers arranged in a square, and a laser distance meter. The extrinsic parameters of the camera are estimated based on the geometric shape of the square projected onto the structure. In order to estimate the extrinsic parameters accurately, Z-axis translation between the camera and the structure is determined using the laser distance meter. Once the extrinsic parameters are estimated, then the scale factor is calculated and used to convert the displacement from pixel coordinates to real-world coordinates. The structural displacement is converted using the calculated scale factor. Note that, in order to measure the displacement from captured video, the ORB algorithm (Rublee et al., 2011) and the Kanade-Lucas-Tomasi (KLT) tracker (Lucas & Kanade, 1981; Shi, 1994; Tomasi & Kanade, 1991) are implemented for feature extraction and feature tracking, respectively. This chapter describes the development of LAVOLUTION system and describes its alignment, its calibration, and the calculation of the scale factor.

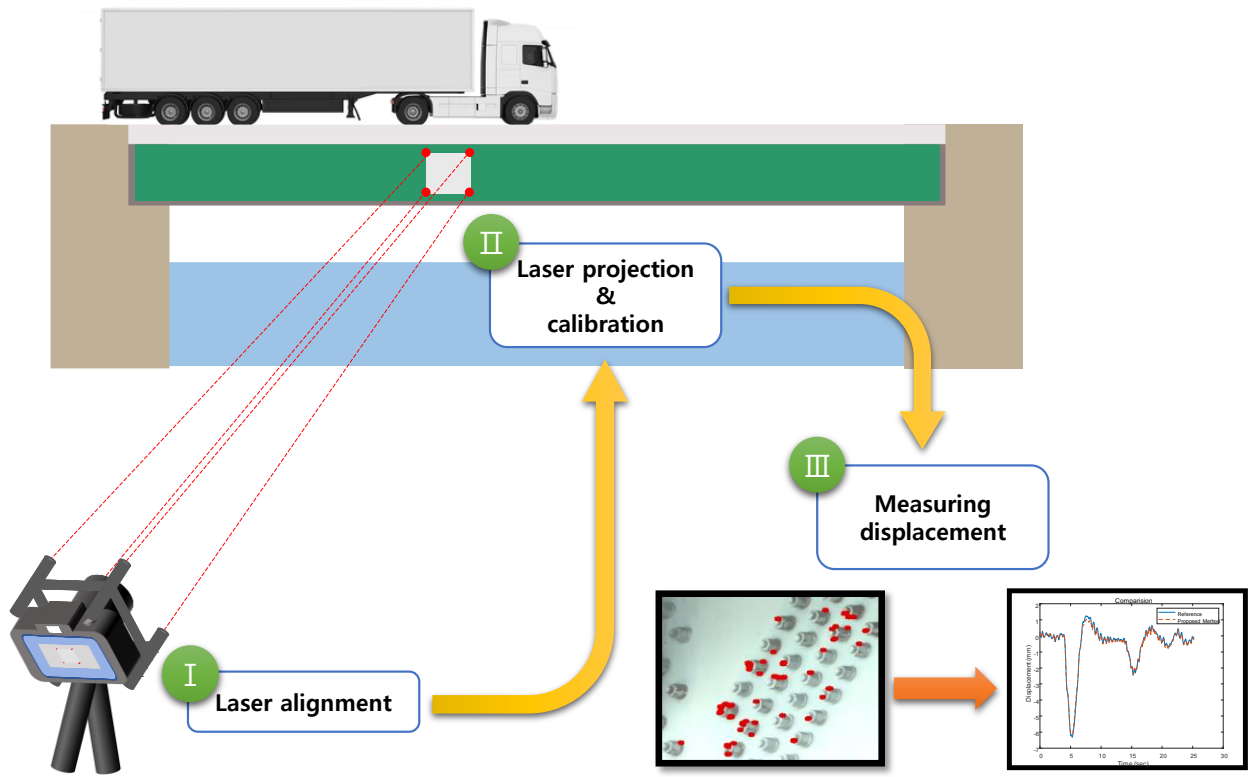

Figure 1. Proposed non-target structural displacement measurement

## 2.2. Development of the jig for tunable structured-light (SL)

In order to calibrate the camera using the shape information from the four laser beams, the lasers need to be aligned parallel to each other. Therefore, the jig has four cylindrical spaces for the laser pointers, as shown in Figure 2. The head of each laser pointer is fixed to the pit at the front of the cylinder using a rubber band and the laser is adjusted using two screws spaced 90 degrees apart (Figure 2b).

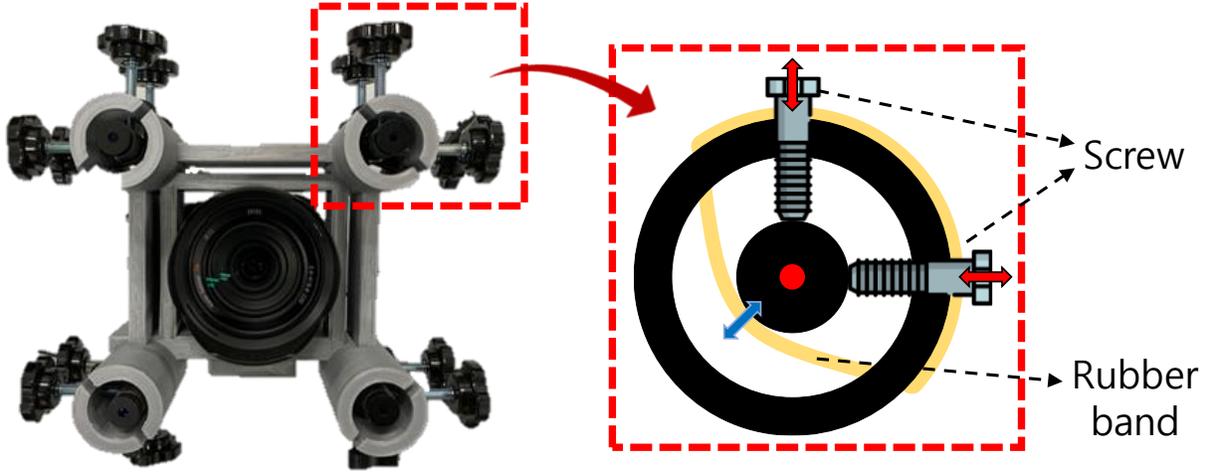

Figure 2. LAVOLUTION jig: (a) body frame of the jig and (b) adjustment of the laser pointers using a rubber band and screws

The jig is designed so that the camera and four laser pointers behave as a rigid body. Thus, the optical axis of the camera also needs to be parallel to the laser pointers. To align them, the laser beams are projected onto a white screen set at a certain distance (Figure 3a). Virtual calibration points arranged in a square are then generated to compare the projected laser points with their ideal location. Each virtual calibration point is located at an ideal position where a laser point should be projected when the optical axis is perpendicular to the screen at the set distance. The virtual calibration points are generated as follows:

$$P_v^i = f(D, K, R, T, T_s) = \lambda K [R|T] \begin{bmatrix} -\frac{T_s}{2} & \frac{T_s}{2} & \frac{T_s}{2} & -\frac{T_s}{2} \\ \frac{T_s}{2} & \frac{T_s}{2} & -\frac{T_s}{2} & -\frac{T_s}{2} \\ D & D & D & D \\ 1 & 1 & 1 & 1 \end{bmatrix}, \quad (1)$$

where $D$ is the distance between the camera and the projection plane (i.e., the white screen), $K$ is the intrinsic matrix, $R$ is the rotation matrix, $T$ is the translation vector, and $T_s$ is the length of the two laser points.

$$K = \begin{bmatrix} f_x & 0 & c_x \\ 0 & f_y & c_y \\ 0 & 0 & 1 \end{bmatrix} \qquad (2)$$

For the alignment of the lasers, the optical axis and screen rotation are assumed to be zero, so the rotation matrix is an identity matrix. Once the virtual calibration points are generated, the laser points can be aligned to match the virtual calibration points by manually adjusting the screws of the jig (Figure 3b).

(a)  (b)

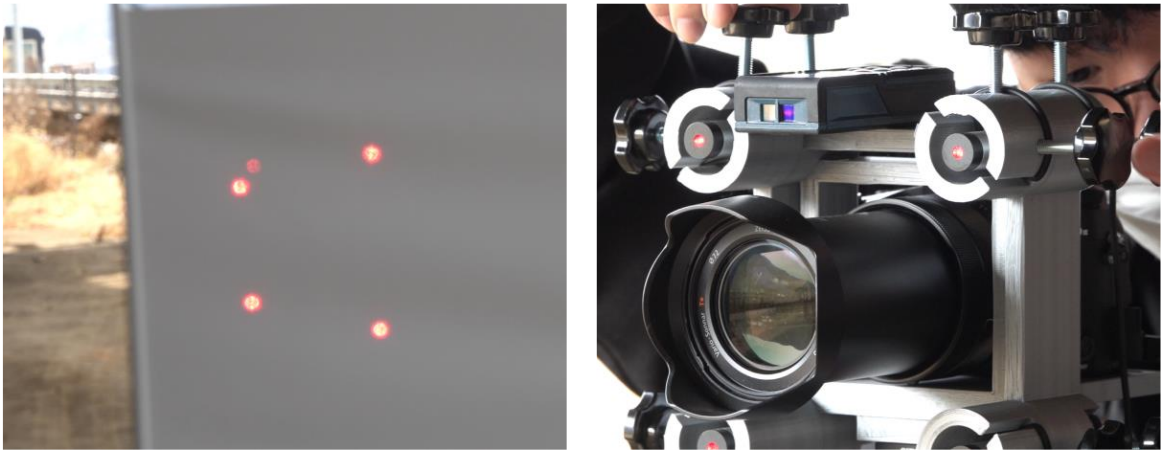

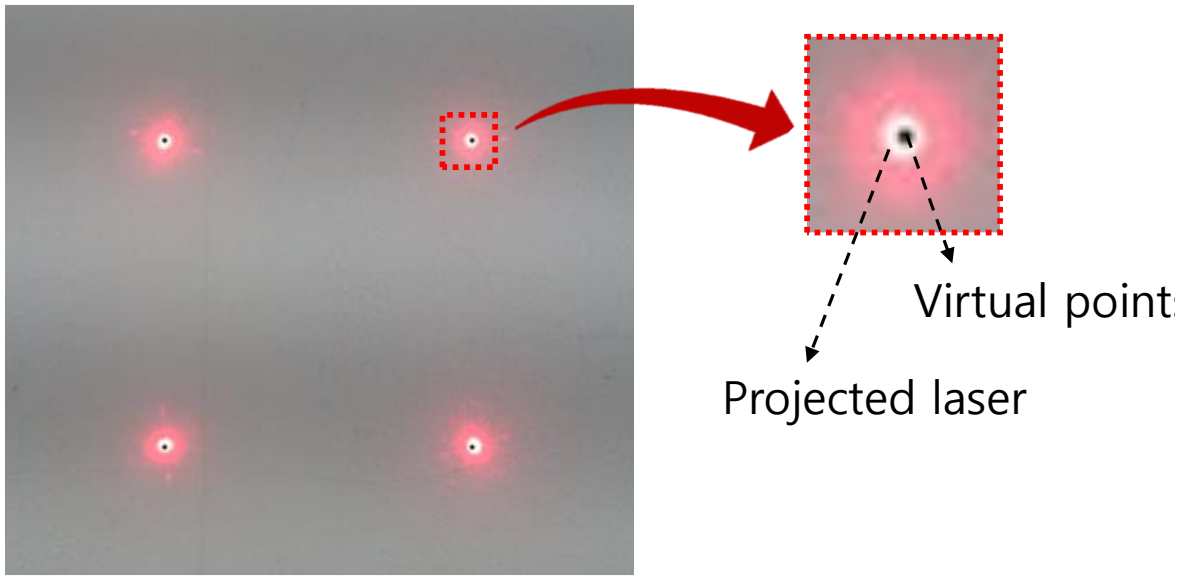

Figure 3. Laser alignment process: (a) projected laser points on the screen, (b) adjustment of the jig, and (c) laser points aligned to the virtual calibration points (black dots).

However, in practice, vertically aligning the optical axis of the camera with a distant screen is difficult without the use of additional devices. If the optical axis and the screen are not perpendicular to each other, the laser pointers will not align properly, which may cause problems in the measurement of the displacement.

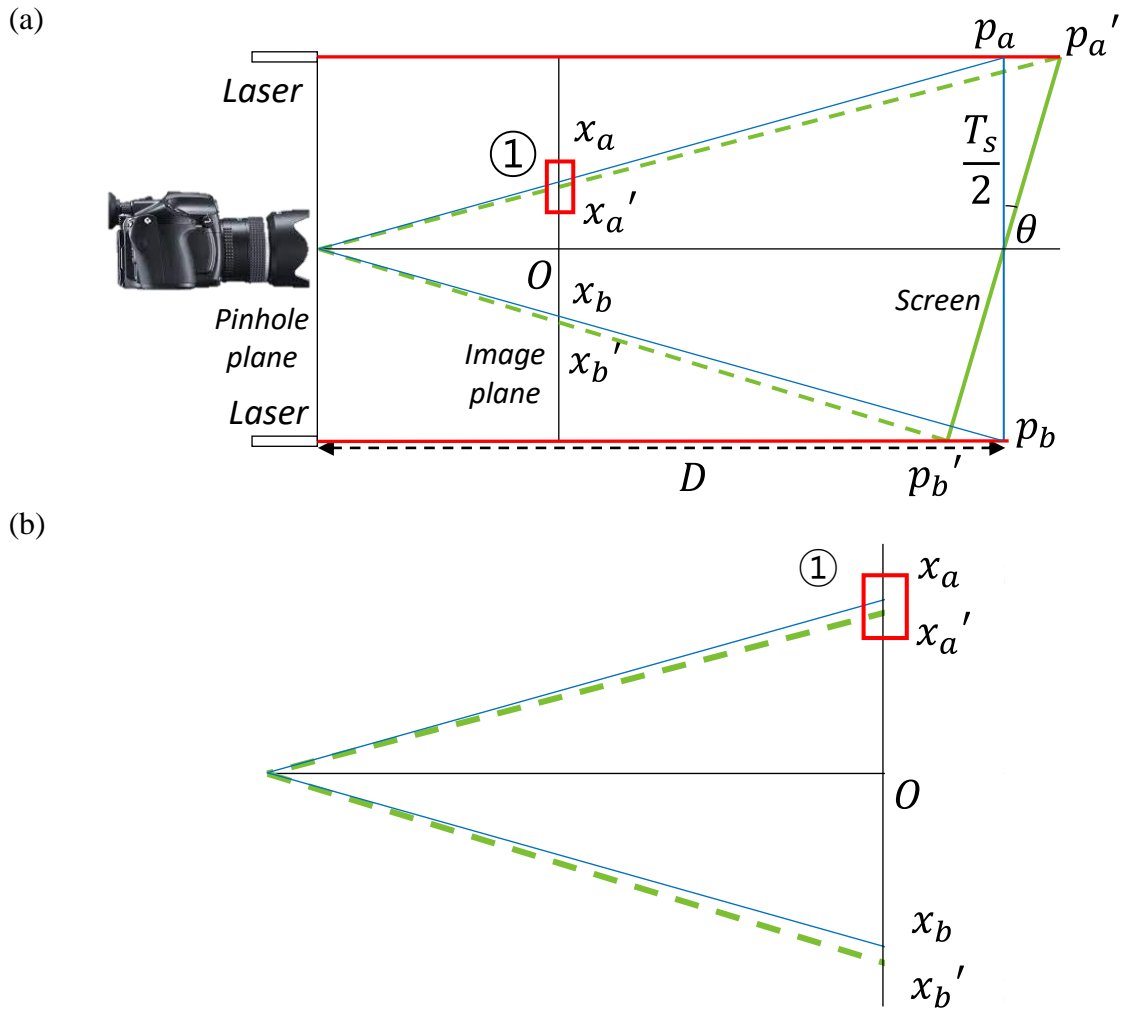

Figure 4. Alignment error on the image plane: (a) perpendicular and tilted planes along the optical axis and (b) the pixel difference for the perpendicular and tilted planes

Figure 4(a) presents two cases, one where the screen is perpendicular to the optical axis and the other where it is tilted by angle $\theta$. The difference in the camera coordinates caused by the tilt is presented in Figure 4(b). The difference can be represented by the ratio of the length between the camera coordinates, which is associated with focal length $f$ and distance $D$. For the perpendicular plane, the ratio of $f$ and $D$ is directly proportional to that of the image and target as follows:

$$\frac{x_a}{f} = \frac{T_s}{2} \bigg/ D$$

$$\frac{x_b}{f} = \frac{T_s}{2} \bigg/ D$$
(3)

On the other hand, for a tilted plane, the ratio is affected by the tilt angle as follows:

$$\frac{x'_a}{f} = \frac{T_s}{2} \bigg/ \left(D - \frac{T_s}{2}\tan\theta\right)$$

$$\frac{x'_b}{f} = \frac{T_s}{2} \bigg/ \left(D + \frac{T_s}{2}\tan\theta\right)$$
(4)

The length of camera coordinates for each plane is thus

$$|x_a + x_b| = \frac{T_s}{D} f$$

$$|x'_a + x'_b| = \frac{f\frac{T_s}{2}\cos\theta}{D\cos\theta - \frac{T_s}{2}\sin\theta} + \frac{f\frac{T_s}{2}\cos\theta}{D\cos\theta + \frac{T_s}{2}\sin\theta} = \frac{DT_s f \cos^2\theta}{\left(\frac{4D^2}{T_s^2} + 1\right)\cos^2\theta - 1}$$
(5)

The ratio of the two lengths in Eq. (5) can be considered as the difference in the camera coordinates; as the distance increases, the ratio converges to 1:

$$\lim_{D \to \infty} R = \frac{|x'_a + x'_b|}{|x_a + x_b|} \approx 1 \tag{6}$$

Therefore, when the laser alignment is conducted using a larger distance, the influence of the angle is reduced, and the alignment is more precise.

A numerical simulation is conducted to validate the lower effect of the angle with respect to the distance. For this simulation, the screen is tilted by a pitch of 10 degrees and a yaw of 10 degrees. Figure 5 shows the position error of the projected laser point as a function of the distance. The position error is the average of the distance moved by the four laser points when the pitch and yaw are zero degrees and tilted by 10 degrees. The positional error decreases with respect to the distance between the jig and a screen. At a distance of 6 m, the positional error is less than 1 pixel. The movement of the laser points caused by 10 degrees of pitch and yaw is very small at a distance of 6 m, thus the influence of the angle can be ignored. Thus, for the precise alignment of the lasers and the camera, the screen should be set at a minimum distance of 6 m.

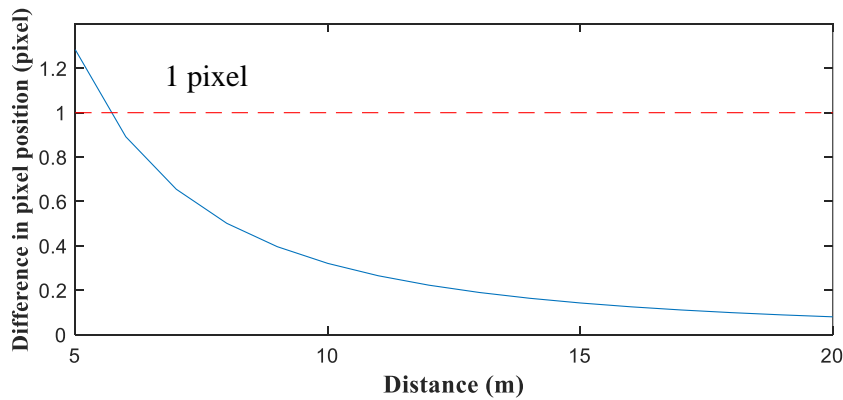

Figure 5. Difference in the pixel position with respect to the distance between LAVOLUTION and the screen

*2.3. Calibration using four laser points*

In order to measure structural displacement using the proposed vision system, camera calibration to estimate the camera parameters and the scale factor is required. Using LAVOLUTION, the four laser points can be projected onto the target structure for calibration and to calculate the scale factor.

Following the alignment process described in Section 2.2, the developed jig is rotated along the x- and y-axes to face the target structure. In order to precisely estimate the structural displacement, the rotation should be determined carefully; however, the initial angle is determined using an inclinometer and, due to sensor errors, the initial angle should be calibrated. The projected laser points and estimated projection points generated for a known initial angle do not perfectly match due to an error in the angle. Therefore, the pitch and yaw are estimated by comparing and optimizing the coordinates of the estimated projection points in relation to the coordinates of the center of each laser point. The coordinates of the projected laser points are detected using the Otsu algorithm (Otsu, 1979), which determines an optimal threshold by representing the intensity distribution of the image pixel points as a histogram. After binarizing the image to the threshold obtained using the Otsu algorithm, the coordinates of the bright region are set as the center coordinates of the laser points. The nonlinear least-squares method is used to optimize the initial angle to estimate the optimal angle by reducing the error between the coordinates of the projected points and the estimated projection points as follows:

$$\min_{R} \sum_{i=1}^{n} \left\| P_p^i - P_e^i \right\|, \tag{7}$$

where $\theta$ and $\phi$ are the pitch and yaw, respectively, $P_p^i$ is the coordinates of the projected points, and $P_e^i$ is the coordinates of the estimated projection points. The estimated projection points are generated as follows:

$$P_e^i = f(D, K, R, T, T_s) = \lambda K [R|T] \begin{bmatrix} -\frac{T_s}{2} & \frac{T_s}{2} & \frac{T_s}{2} & -\frac{T_s}{2} \\ \frac{T_s}{2} & \frac{T_s}{2} & -\frac{T_s}{2} & -\frac{T_s}{2} \\ D & D & D & D \\ 1 & 1 & 1 & 1 \end{bmatrix}, \tag{8}$$

where translation vector $T$ is a zero vector and rotation matrix $R$ is not an identity matrix, but a function of the pitch and yaw as follows:

$$R = g(0, \varphi, \theta) = \begin{bmatrix} \cos\varphi & 0 & \sin\varphi \\ \sin\theta\sin\varphi & \cos\theta & -\sin\theta\cos\varphi \\ -\cos\theta\sin\varphi & \sin\theta & \cos\theta\cos\varphi \end{bmatrix}, \tag{9}$$

where the roll is zero, $\varphi$ is the yaw, and $\theta$ is the pitch. The estimated projection points differ from the virtual calibration points in Eq. (1) because the rotation matrix is not an identity matrix due to camera rotation. Once optimization using nonlinear least squares is complete, the estimated pitch and yaw are employed as the actual angle of the camera and the surface.

*2.4. Scale factor calculation*

The final calibration step is to calculate the scale factor to link the pixels to the corresponding distance in real-world coordinates. As shown in Figure 4, the laser points are projected at the same vertical distance along the optical axis regardless of whether the object plane is rotated or not. When the image plane is parallel to the object plane, the scale factor can be measured as

$$SF = T_s / \overline{x_a x_b}, \tag{10}$$

where $T_s$ is the target size based on the laser points of the developed jig, and $\overline{x_a x_b}$ is the pixel distance on the image plane.

When the object plane is not parallel to the image plane, i.e., the camera is rotated, $P'_a$ and $P'_b$ are the projected laser points. Laser beams are straight, so the points are projected at the same global distance perpendicular to the optical axis regardless of whether the object plane is rotated or not. From another perspective, the actual distance between the laser points projected on a tilted object plane increases with rotation angle $\theta$. Therefore, the rotation angle should be considered when calculating the scale factor. Consequently, the scale factor for structural displacement estimation can be calculated as

$$SF = T_s / \overline{x'_a x'_b} / \cos\theta, \tag{11}$$

where $\theta$ is the pitch. Note that only the vertical displacement of the structure is considered, so the yaw does not affect the scale factor.

In order to verify the effectiveness of the proposed scale factor, a simulation is conducted to demonstrate the effect of increasing the pitch and yaw. The error of the scale factor is calculated by comparing the scale factor of the target with that of the proposed method as follows:

$$Error = \frac{1}{N}\sum_{i=1}^{N}\left(\frac{SF_{target}^{i} - SF_{proposed}^{i}}{SF_{target}^{i}}\right) \qquad (12)$$

Note that, for each pitch and yaw, the error of the scale factor is calculated by averaging 100 simulations, so $N$ is 100 in this case.

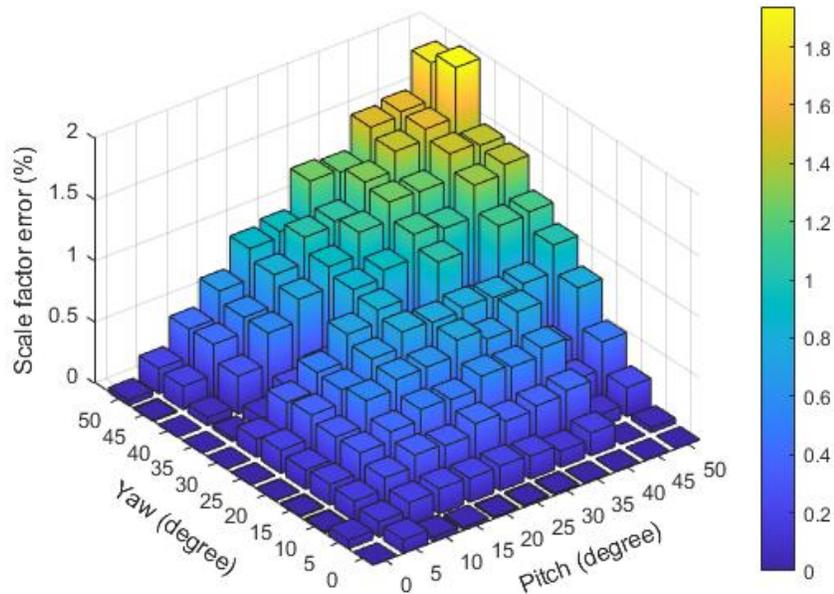

Figure 6. Scale factor error with respect to the pitch and yaw

Table 1. Scale factor error for the target-based and proposed calculations

| Scale factor error (%) | | Pitch | | | | | |
|---|---|---|---|---|---|---|---|
| | | 0° | 10° | 20° | 30° | 40° | 50° |
| Yaw | 0° | 0 | 0.04 | 0.01 | 0.02 | 0 | 0 |
| | 10° | 0.10 | 0.26 | 0.33 | 0.34 | 0.40 | 0.24 |
| | 20° | 0 | 0.33 | 0.61 | 0.74 | 0.78 | 0.79 |
| | 30° | 0 | 0.35 | 0.72 | 0.11 | 0.57 | 1.15 |
| | 40° | 0 | 0.31 | 0.75 | 1.10 | 1.25 | 1.43 |
| | 50° | 0.04 | 0.43 | 0.90 | 1.26 | 1.51 | 1.85 |

Figure 6 and Table 1 show that, if the angle is less than 50 degrees, the error between the scale factor calculated from the target-based method and the proposed method is less than 2%. In most cases in the field, the camera is rarely rotated more than 30 degrees to capture the target structure, hence the structural displacement can be accurately measured with the proposed system.

## 3. Lab-Scale Experimental Validation

### 3.1. Validation of proposed scale factor calculation

A lab-scale experiment is conducted to validate the proposed scale factor calculation. Figure 7a shows the experimental setup, with a total of nine markers attached to produce rotations of 0, 10, and 30 degrees in terms of the pitch and yaw. ArUco markers (Figure 7b) are used to calculate the reference scale factor to compare with the proposed scale factor calculation. The reference scale factor using the ArUco markers is calculated using the OpenCV function (Bradski & Kaehler, 2015). Table 2 shows the error between the proposed scale factor calculation and the true scale factor calculated using the target-based method. The experimental results show that the maximum error is less than 3%, validating the numerical simulations.

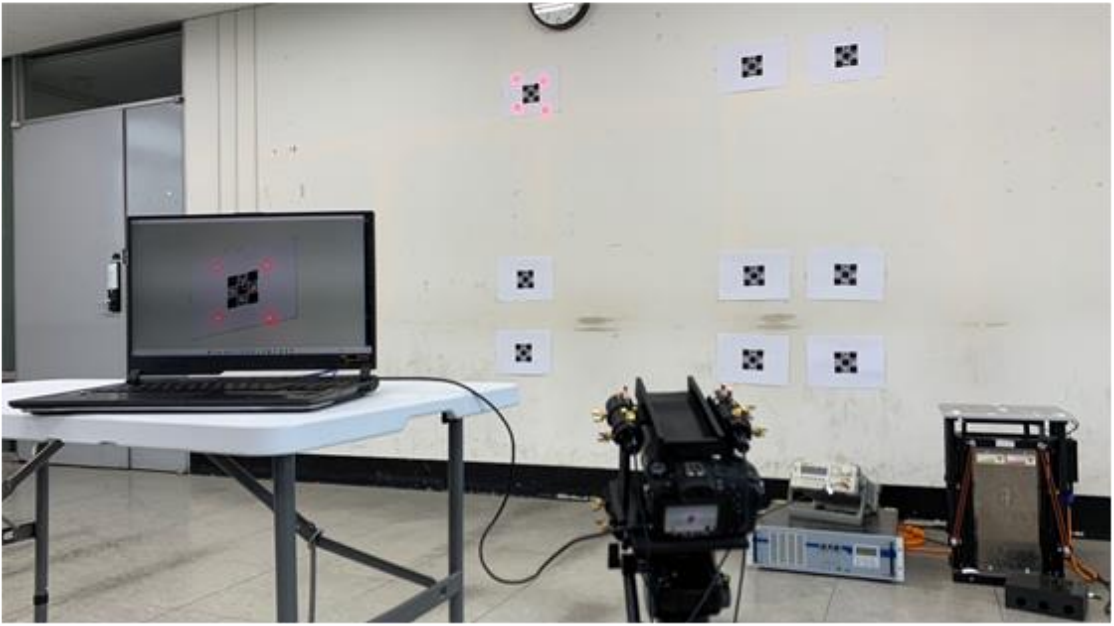

(a)

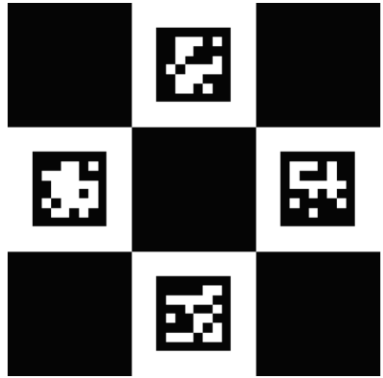

(b)

Figure 7. Experimental setup for the validation of the proposed scale factor calculation: (a) scale factor calculation and (b) reference ArUco marker

Table 2. Scale factor error between the target-based and proposed calculations

| Scale factor error (%) | | Pitch | | |
|---|---|---|---|---|
| | | 0° | 10° | 30° |
| Yaw | 0° | 0.82 | 0.97 | 1.56 |
| | 10° | 1.48 | 1.61 | 2.27 |
| | 30° | 1.75 | 1.95 | 2.69 |

*3.2. Validation of the proposed displacement estimation using a shaking table*

A series of experiments is used to validate the proposed method and to compare it with a reference LDV. An overview of the experimental setup is described in Figure 8. In the experiment, a shaking table is used as a testbed. Video is taken with a Sony RX10 M3 camera 1.5 m away from the shaking table using a 2K FHD (30 fps, 1920 × 1080-pixel resolution) setting. The reference displacement is measured using an ILD-1420 with a 1 kHz sampling rate. Two sets of experiments are conducted by rotating the proposed LAVOLUTION.

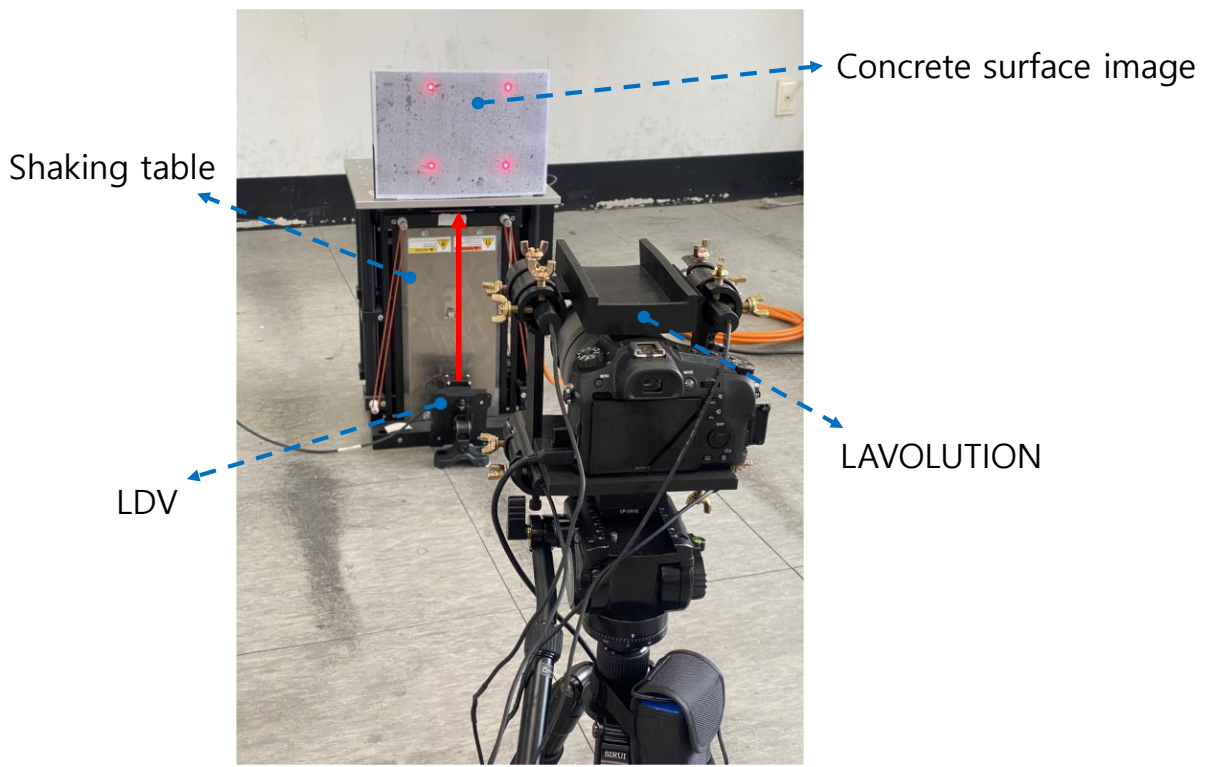

Figure 8. Experimental setup for the comparison of the proposed method with an LDV

The measured displacement using the proposed method and the reference LDV are shown in Figure 9, and the maximum displacement is compared in Table 3. The proposed method has a maximum displacement error of 1.31% and 1.77%, respectively, compared to the reference LDV. Comparing the differences in the maximum displacement in both cases, the proposed method has a small error of 0.07 mm and 0.08 mm, respectively, validating its accuracy in measuring dynamic displacements of less than 0.1 mm.

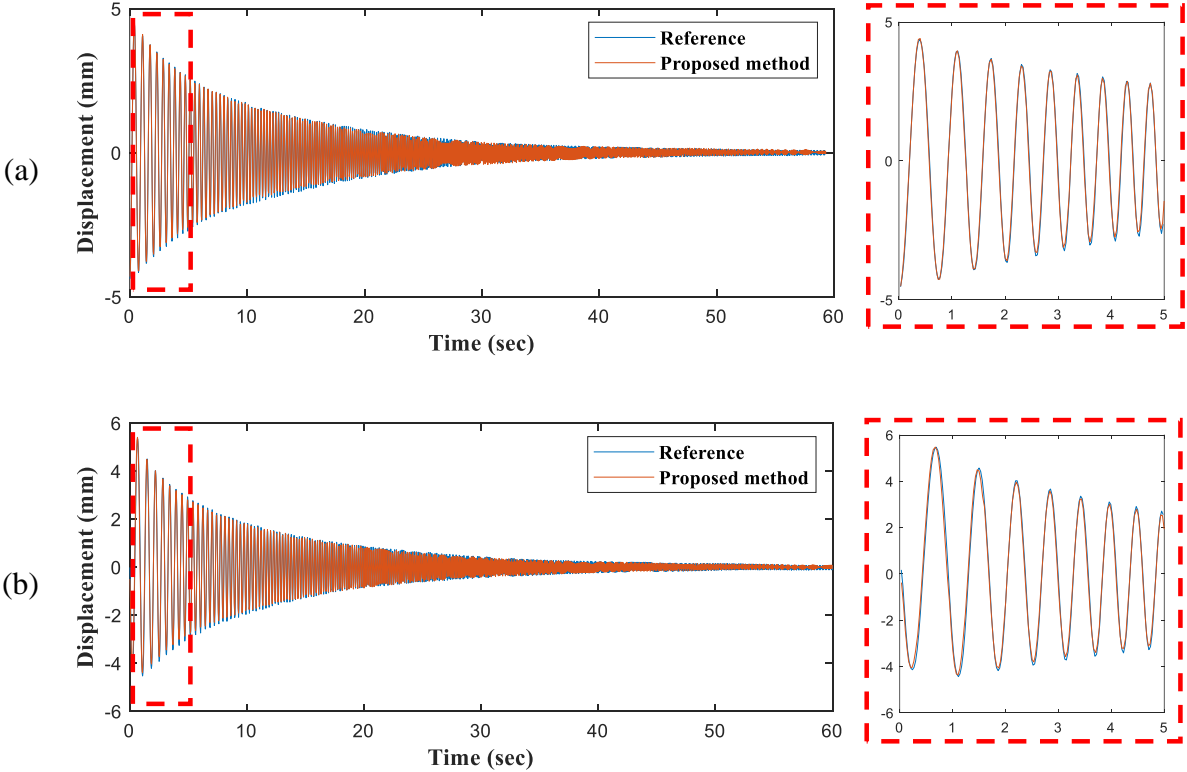

Figure 9. Comparison of the measured structural displacement of shaking table: (a) case 1 – pitch of 0 degrees and yaw of 18 degrees, and (b) case 2 – pitch of 16 degrees and yaw of 30 degrees

Table 3. Comparison of the displacement measurements for the shaking table

| Case | Method | Maximum Displacement (mm) | Error (mm) | Error (%) |
|---|---|---|---|---|
| Case 1 | Proposed method | 5.41 | 0.07 | 1.31 |
|  | Reference displacement | 5.34 | - | - |
| Case 2 | Proposed method | 4.59 | 0.08 | 1.77 |
|  | Reference displacement | 4.51 | - | - |

## 4. Field Test on a Steel-box Girder Bridge

### 4.1. Experimental setup

An experiment is performed on a steel-box girder bridge located at Namyangju, South Korea. An overview of the experimental setup is presented in Figure 10. Because the proposed method is a non-target-based system, feature points are extracted from the stud bolts of the structure. An LDV is used as a reference for the comparison of the displacement estimation results. The lasers are aligned at a distance of 6 m as proposed in Section 2.2 and camera rotation has a pitch of 35 degrees and a yaw of 5 degrees.

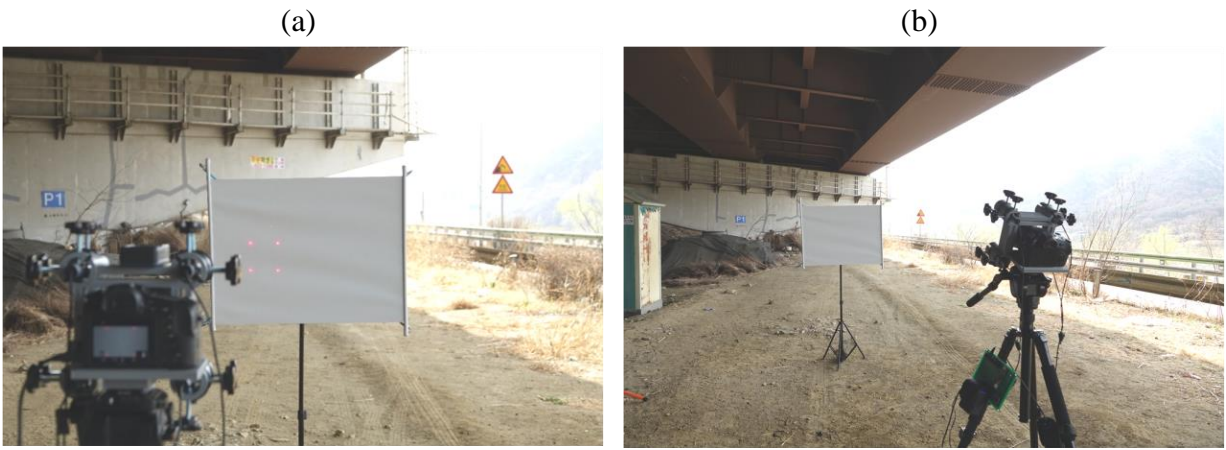

Figure 10. Experimental setup: (a) laser alignment and (b) displacement estimation

The structural displacement is measured via video; Figure 11 presents two examples of the measured displacement compared to the reference LDV. The maximum displacement estimated using the proposed method is 6.17 mm and 6.88 for the two examples. Compared with the 6.33 mm and 6.93 mm measured by the reference LDV, the maximum displacement error is 2.53% and 0.72%, respectively, less than 3% and consistent with the lab-scale experimental results. As shown in Figure 12, the magnitude of the power spectral density (PSD) for the proposed method agrees well with the reference LDV in both examples, indicating that the displacement is accurately measured using the proposed system.

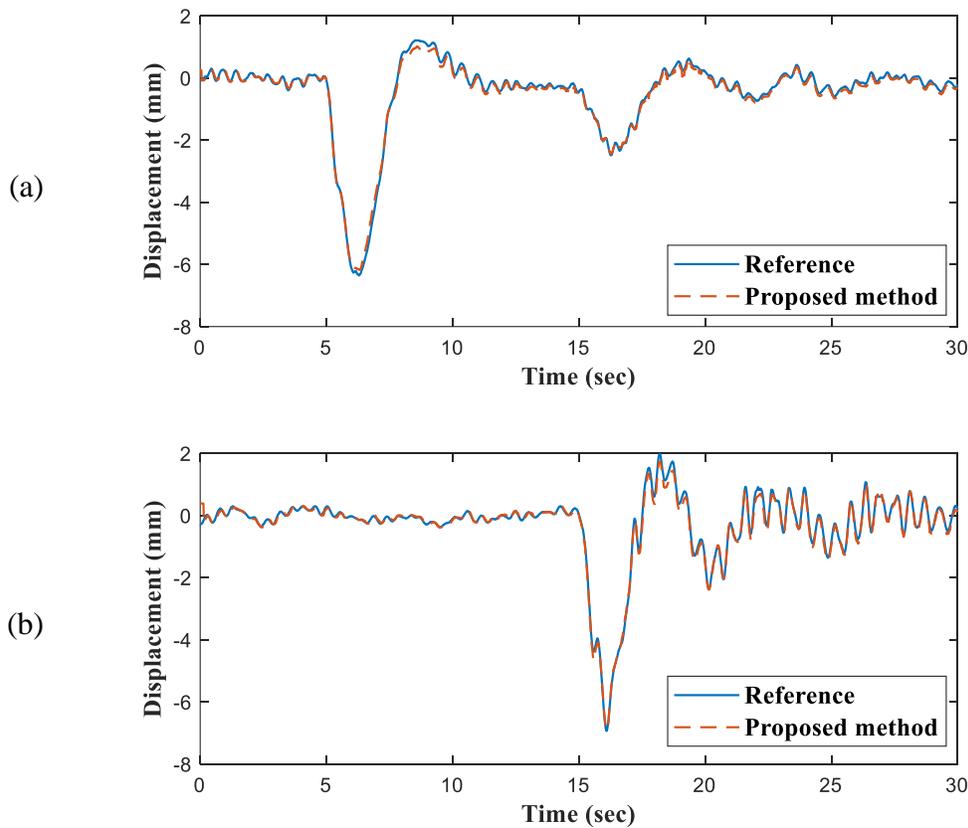

Figure 11. Comparison of the measured structural displacement of steel-box girder bridge

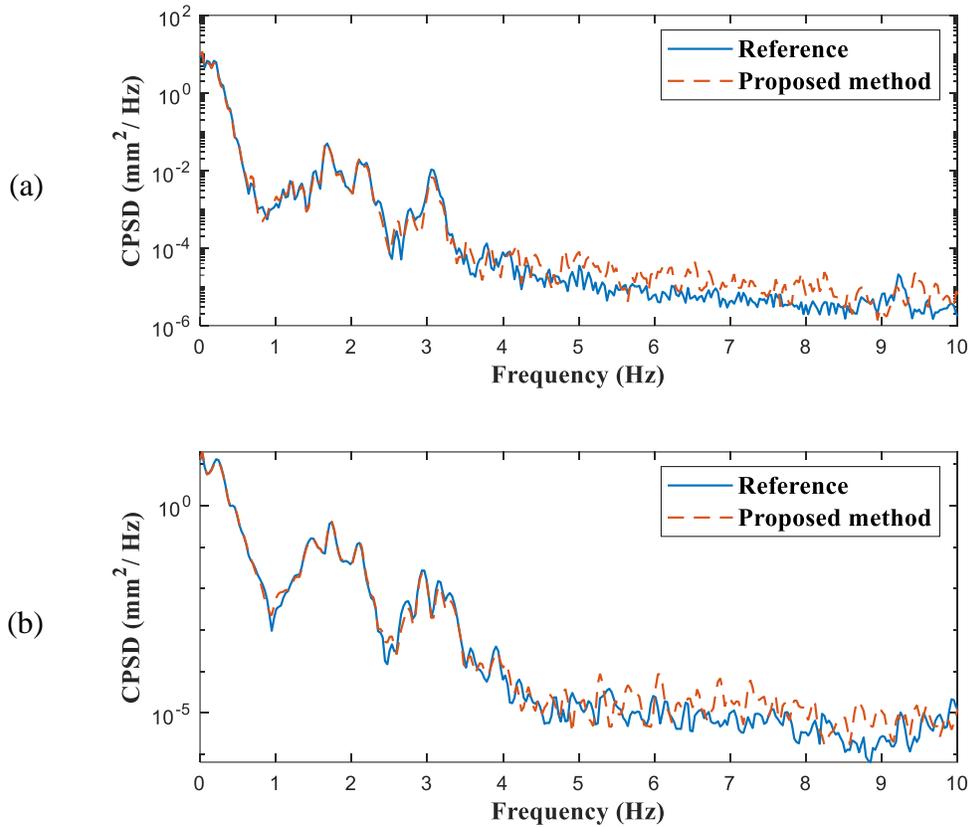

Figure 12. Comparison of the power spectral density of measured structural displacement

## 5. Conclusions

This paper presents a non-target structural displacement measurement strategy that makes use of SL and CV. A camera integrated with tunable SL is developed to correct positional errors in the four laser pointers. The laser pointers are calibrated by generating virtual calibration points that mark the exact points where the SL should be located when projected parallel to the target surface. Numerical analysis is employed to determine the distance and angle of rotation between the target surface and the camera. The optimal distance for tuning the SL is determined to be 6 m with an allowable initial rotation error of ±10°.

A scale calibration method is also proposed to determine the camera pose relative to the target surface by minimizing the difference in the position of the projected SL and the analytic model by adjusting the parameters related to the camera pose. Based on the camera pose, the scale factor is calculated, and numerical simulations showed a maximum scale factor error of 1.85% when the relative yaw and pitch are in the range of 0−50°. To validate the proposed system, a lab-scale experiment is conducted carried out. The scale factor obtained from the proposed method is compared with that acquired by markers whose geometry is already known. When tested in different positions for scale factor estimation, the proposed system produced an error of less than 3% compared with a conventional marker-based system. Subsequently, an experiment with a shaking table is used to characterize the accuracy and dynamic range of the system; the proposed method has a maximum displacement error of less than 1.77% for data acquisition at 30 Hz. Finally, a field test is conducted on a steel-box girder bridge and to measure ambient displacement under truck loads. For two loading examples, the error in the maximum displacement is 2.53% and 0.72%, thus confirming the accuracy and effectiveness of the proposed system.

## Funding

This work was supported by a Korea Agency for Infrastructure Technology Advancement (KAIA) grant funded by the Ministry of Land, Infrastructure and Transport (Grant 22CTAP-C164184-02)